\documentclass[letterpaper]{article}
\usepackage{proceed2e}
\usepackage[margin=1in]{geometry}

\usepackage{natbib}

\usepackage{times}

\usepackage{microtype}
\usepackage{graphicx}
\usepackage{booktabs} 

\usepackage{graphics} 
\usepackage{amsmath} 
\usepackage{amssymb}  

\usepackage[caption = false]{subfig}

\usepackage{mathtools}
\usepackage{color}
\usepackage[linkcolor=true,citecolor=blue,allcolors=blue]{hyperref} 
\usepackage{algorithm}
\usepackage{algpseudocode}

\usepackage{booktabs}

\usepackage{todonotes}

\def\mathbi#1{\textbf{\em #1}}

\usepackage{url}

\title{Adaptive Configuration Oracle for Online Portfolio Selection Methods}

\author{} 

\author{ {\bf Favour M.~Nyikosa\thanks{For correspondence, email:\texttt{favour@robots.ox.ac.uk}}} \\
Machine Learning Group\\ 
Oxford-Man Institute\\
University of Oxford\\
\texttt{favour@robots.ox.ac.uk}\\
\And
{\bf Michael A. Osborne}  \\
Machine Learning Group\\ 
Oxford-Man Institute\\
University of Oxford\\
\texttt{mosb@robots.ox.ac.uk}\\
\And
{\bf Stephen J. Roberts}   \\
Machine Learning Group\\ 
Oxford-Man Institute\\
University of Oxford\\
\texttt{sjrob@robots.ox.ac.uk}\\
}

\begin{document}

\maketitle

\begin{abstract}

Financial markets are complex environments that produce enormous amounts of noisy and non-stationary data. One fundamental problem is online portfolio selection, the goal of which is to exploit this data to sequentially select portfolios of assets to achieve positive investment outcomes while managing risks. Various algorithms have been proposed for solving this problem in fields such as finance, statistics and machine learning, among others. Most of the methods have parameters that are estimated from backtests for good performance. Since these algorithms operate on non-stationary data that reflects the complexity of financial markets, we posit that adaptively tuning these parameters in an intelligent manner is a remedy for dealing with this complexity. In this paper, we model the mapping between the parameter space and the space of performance metrics using a Gaussian process prior. We then propose an oracle based on adaptive Bayesian optimization for automatically and adaptively configuring online portfolio selection methods. We test the efficacy of our solution on algorithms operating on equity and index data from various markets. 

\end{abstract}

\section{INTRODUCTION}

There has been a resurgence of interest in portfolio choice problems over the last decade due to the availability of large amounts of financial data and cheap computing power. The empirical evidence from the accumulated data has motivated trading strategies that attempt to exploit the behaviour of predictable components in the data. One variant of portfolio selection that has become increasingly popular is the online portfolio selection (OLPS) problem, where one seeks to sequentially select portfolios over a fixed time horizon to maximize some investor-specified criterion. For a detailed survey of OLPS, see \cite{fin15}.

Most online portfolio selection methods estimate their parameters from backtests on data, or by hand-tuning. These parameters are then used for the rest of the out-of-sample trading period. However, there are some limitations of using this approach for parameter estimation.

First, the methods used to estimate parameters tend to involve a lot of human intervention and are often time-consuming. For cases where the algorithm is particularly sensitive to the parameters, the configuration activity becomes a critical part of determining whether a method will succeed, often overriding the importance of the underlying assumptions of the strategy.

Secondly, financial data is often non-stationary, and because online portfolio selection algorithms operate sequentially, the non-stationarity in the data affects the efficiency of the methods. An algorithm configured on backtested data may start out doing well on out-of-sample scenarios, but its performance may deteriorate after several trading periods. 

Lastly, because trading on a financial market is a costly endeavour, the non-stationarity in the market data that affects the trading efficacy introduces new risks relating to the parameters used for the portfolio selection method. Since these risks are not managed directly or adequately exploited in many of the existing approaches, they can negatively affect the trading algorithm's performance. 

We address these issues by utilizing a Bayesian nonparametric approach to modelling a relationship between the parameters and the performance criterion in every period using a spatiotemporal Gaussian process (GP) prior. We then propose a mechanism for using adaptive Bayesian optimization for tuning the parameters of online portfolio selection algorithms. The technique enables the methods to adapt to changing market conditions by having a handle on the uncertainty of the model, parameters and data. We consider a range of investment strategies driven by widely accepted trading characteristics. Our formulation thus provides an oracle for tuning existing algorithms and has the advantage of delivering interpretable performance improvements. This interpretability offers useful algorithmic insights to practitioners.

This paper is structured as follows. Section \ref{sec: olps-background} highlights some background and Section \ref{sec: olps-problem} outlines the problem setting. Section \ref{sec: olps-solution} describes the adaptive algorithm configuration oracle. Lastly, Section \ref{sec: olps-experiments} includes experiments and Section \ref{sec: olps-discussion} the discussions.

\section{BACKGROUND}\label{sec: olps-background}

\subsection{BAYESIAN OPTIMIZATION}\label{sec: olps-bo}

Let us consider an optimization problem
	\begin{equation}\label{eq: olps-bo-def}
			\mathbi{x}^{\star} = \underset{ \mathbi{x} \in \mathcal{X} }{ \text{arg min} }f(\mathbi{x}),
	\end{equation}
	where $\mathbi{x} \in \mathcal{X} \subset \mathbb{R}^{D}$ and $f$ is unknown, noisy and expensive to evaluate. Bayesian optimization (BO) is a sequential design strategy for optimizing a function of this nature. BO uses a surrogate model to learn the latent objective function from available samples and searches for a suitable point to sample the objective function. This search is performed using some predefined heuristics to get closer to the optimum.  The general strategy is to have heuristics that intelligently and automatically \emph{explore} and \emph{exploit} the objective function $f(\mathbi{x})$. In practice, this sampling choice is achieved by performing a secondary optimization of a surrogate-dependent acquisition function $a( \mathbi{x} )$. Popular acquisition functions that are often used in practice are the expected improvement (EI) \citep{bo7} and upper confidence bounds (UCB) \citep{bo23}. After the proposal input point is obtained, the objective function is evaluated at that point, and the process is repeated. 
	For a detailed survey of Bayesian optimization, see \cite{bo25}.
%
%

\subsection{GAUSSIAN PROCESSES}\label{sec: olps-gp}

A popular surrogate model for Bayesian optimization is a Gaussian process (GP) prior. A GP $\mathcal{GP}(\mu, K)$ is a collection of random variables where any finite subset has a joint Gaussian distribution. It describes a prior distribution over functions, and it is completely specified by a mean function 
	\begin{align}\notag
		&\mu: \mathcal{X}  \mapsto \mathbb{R}, \\ 
		&\mu(\mathbi{x})  = \mathbb{E}\big[ f(\mathbi{x})  \big],
	\end{align} 
	and covariance function (or kernel) 
	\begin{align} \notag
		&K: \mathcal{X} \times \mathcal{X}  \mapsto \mathbb{R}, \\ 
		&K( \mathbi{x}, \mathbi{x}^{\prime} )  = \mathbb{E} \big[ ( f(\mathbi{x}) - \mu(\mathbi{x})) ( f( \mathbi{x}^{\prime} ) - \mu(\mathbi{x}^{\prime} ))  \big],
	\end{align}
	where  $K(\cdot, \cdot) \leq 1$. For a detailed overview, see \cite{gp1}.


\subsection{ADAPTIVE BAYESIAN OPTIMIZATION}\label{sec: olps-bo}

Adaptive Bayesian optimization \citep{nyikosa2018ABO} extends standard Bayesian optimization to solve expensive and latent dynamic optimization problems. A dynamic optimization problem (DOP) \citep{dop1} is regarded to be:
	\begin{flalign}\label{eq: dop_intro}
	    \text{DOP } = \Big \{ 
	    \begin{split} 
	        \text{ minimize      }  &    f( \mathbi{x} , t)\\
	        \text{ s.t. \: \: \: \:   }  &    \{ \mathbi{x}, t \} \in F(t) \subseteq \mathcal{S}, t \in \mathcal{T} \:                                  
	    \end{split}
	     \Big \}
	\end{flalign}
	where:
	\begin{itemize}
	   \item $\mathcal{S} \in \mathbb{R}^{D}$, $\mathcal{S}$ is the search space. 
	   \item $t \in \mathbb{R}$ is the time.
	   \item $ f: \mathcal{S} \times \mathcal{T}  \mapsto \mathbb{R} $ is the objective function that assigns a numerical value ($f(\mathbi{x}, t) \in \mathbb{R}$) to each possible solution ($ \mathbi{x} \in \mathcal{S} $) at time $t$.
	   \item $F(t)$ is the set of all feasible solutions $ \{ \mathbi{x}, t \} \in F(t) \subseteq \mathcal{S} $ at time $t$.
	\end{itemize}

A DOP is characterized by a situation where the objective function, or its  constraints, change with time. The solution of a DOP aims to efficiently keep track of the movement of the minimum through the solution space within a fixed and pre-defined horizon of iterations. It is assumed that the temporal evolution is an inherent property of the DOP such that the evaluations of the optimization algorithm do not modify the DOP.

BO is extended in two ways to deal with DOPs. First, the DOP is modelled using a spatiotemporal GP prior with a separable kernel of the form
	\begin{equation}\label{eq: olps-abo-kernel}
			K(f(\mathbi{x}, t), f(\mathbi{x}^{\prime}, t^{\prime})) = K_{\text{S}}({\mathbi{x}, \mathbi{x}^{\prime} })  \odot K_{\text{T}}({t}, t^{\prime}), 
	\end{equation}
	where $K_{\text{S}}(\cdot,\cdot)$ and $K_{\text{T}}(\cdot,\cdot)$ are the spatial and temporal covariance functions, respectively, and $\odot$ is an element-wise or Hadamard product.
	
Secondly, temporal length-scale information from the trained spatiotemporal GP model is used to set the bounds of the feasible region $F(t)$ in every iteration. The bounds are set such that the feasible region represents the part of $f$'s domain where the Gaussian process model is most informative in forecasting future changes relative to the current time $t$. This region is defined by a set of box constraints on both the temporal and spatial variables for the acquisition function optimization. This optimization determines the optimal time $t$, which is a short temporal distance into the future, to evaluate the function at $\mathbi{x}$. This feasible region is adjusted in every iteration based on the Bayesian updates from previous steps. Consequently, the algorithm determines \emph{where} and \emph{when} to evaluate $f$ to induce the tracking behaviour. Therefore, performing the search in this region keeps track of a temporally evolving minimum. 
%
%
%
%
%
%
%
%
%
	
The learnt temporal length-scales for the GP model are an informative tool for determining if temporal changes in $f$ are taking place. For example, very short temporal length-scales mean that there are fast temporal changes in $f$ while very long temporal length-scales imply there are no changes over time.

In practice, there are some circumstances where the times when to make evaluations of $f$ are known in advance, most commonly in the form of how frequently the function is evaluated. In such cases, the temporal part of the box constraints for the feasible region $F(t)$ is set to be the known time of interest, vastly simplifying the problem.

The acquisition function used for ABO determines the type of tracking behaviour induced by the algorithm. Since most acquisition functions aim to explore and exploit the objective function automatically, some of the explore steps lead to suboptimal evaluations. However, these suboptimal steps are beneficial to future iterations because they provide valuable information for the Bayesian updates.
	
When using the UCB acquisition function, adaptive Bayesian optimization becomes a special case of contextual Gaussian process bandit optimization, where the context under consideration is time. For more details on the theoretical analyses of this characterization of Bayesian optimization, see \cite{krause2011contextual} and \cite{bo28}.     

\section{PROBLEM SETTING}\label{sec: olps-problem}

Let us consider an investment task over a complex financial market with $m$ assets for $T$ investment periods. In each period, the asset prices change, and this affects any investment in those assets by a factor proportional to the price changes (also called \emph{returns}). 

We describe an investment decision in every trading period using a \emph{portfolio} of assets, mathematically specified by a portfolio vector $\mathbi{w} \in \mathbb{R}^{m}$, whose elements represent the proportion of wealth invested in each asset. We assume that a portfolio's investment is self-financed such that no margin or short-selling is allowed, i.e. there is no money market. So $\mathbi{w}$ is vector such that $\sum_{i=1}^{m} w_i = 1$ and $w_i \geq 0$.

An investment strategy is represented by a procedure that produces a sequence of portfolios, one for each period, over the horizon of trading periods. The strategy reinvests the portfolio after every period over the horizon.
After $T$ trading periods, a strategy is evaluated by how much money it has made or lost, how well it managed risk and how resilient it was to market fluctuations. This procedure is shown in Algorithm \ref{alg: olps-general}.
	\begin{algorithm}
		\caption{Online Portfolio Selection (OLPS) Problem}
		\label{alg: olps-general}
		\begin{algorithmic}[1]		
			\For {$t = 1,2, ..., T$}
				\State \textcolor{blue}{Strategy learns portfolio $\mathbi{w}_t^{\star}$}
				\State Market reveals asset returns
				\State Portfolio $\mathbi{w}_{t }^{  \star  }$ incurs period profits or losses
				\State Strategy updates portfolio selection model 
			\EndFor
		\end{algorithmic}
	\end{algorithm}

We are concerned with the step that calculates the best portfolio for period $t$ highlighted in Line 2, marked in blue, which we will define as
	\begin{equation}\label{eq: olps-strategy}
			\mathbi{w}_{t }^{  \star  } = \text{ arg max } \; g( \mathbi{w}_{t} ; \mathcal{D},  \boldsymbol{ \theta } ),
	\end{equation}
	where $ \mathbi{w}_{t}$ is the portfolio at trading period $t$, $\mathcal{D}$ is a collation of all the market data and information from previous investment decisions, and $\boldsymbol{ \theta }$ is a parameter vector for the strategy criterion function $g$.
	
The function $g$ defines trading strategies that use market characteristics to exploit market conditions. These investment strategies are based on well-understood market phenomena and empirical observations. The strategies described by $g$ depend on data $\mathcal{D}$ assembled from a myriad of sources, selected for how informative they are for unveiling exploitable trading opportunities. These strategies often have parameters $\boldsymbol{\theta}$ resulting from the construction of the method or theoretical analyses. These parameters often dictate the mode or degree to which certain strategy-centric operations are performed.  Examples of these parameters include, but are not limited to, learning rates, regularization parameters, window sizes, scaling multipliers, tolerances and heuristic parameters. 


The optimization of strategy $g$ shown in Equation \ref{eq: olps-strategy} is usually done using standard off-the-shelf optimizers. This optimization seeks to use the rules of the strategy to find the best portfolio at time $t$ when data $\mathcal{D}$ is encountered, all under the strategic context described by parameters $\boldsymbol{\theta}$. These parameters are usually estimated from backtests or diligently hand-tuned, both done to select parameters with the best performance on past training data before the strategy is executed out-of-sample. These estimated parameters are then used for the optimization in Equation \ref{eq: olps-strategy} in all the future trading periods.     
However, as shown by Algorithm \ref{alg: olps-general}, noisy and non-stationary market information is sequentially revealed to the strategy in every out-of-sample trading period. The problem that arises is that the conditions and assumptions that held true on the backtest data on which $\boldsymbol{\theta}$ was estimated on do not always hold for future time periods. This discrepancy can negatively affect the algorithm's performance. 

We assert that a suitable remedy for this predicament is to adaptively tune $\boldsymbol{\theta}$ in an online manner as it encounters new data. This configuration should be done with the goal of maximizing the utility of using a strategy $g$ for selecting portfolios in each trading period. A suitable criterion for this utility is the unseen future return of the portfolio estimated by $g$. The difficulty with doing this is that the objective function that describes this scheme is: (i) unknown: because we have no analytical form of the function and it depends on unseen quantities whose relationship we have not yet discerned; (ii) expensive to evaluate: because one has to trade on a financial market and incur some cost after deciding what parameters to use to obtain the corresponding utility; and (iii) noisy: as a consequence of being dependent on financial data. 		
	
In this chapter, we propose a suitable model for learning this objective function, and a mechanism for keeping track of the best parameters that maximize the utility of using a strategy $g$ in all the trading periods. The main advantage of solving this problem is that it would provide an intelligent and effective method for making a strategy $g$ acclimatise to different information regimes as it encounters new data.

\section{ADAPTIVE CONFIGURATION ORACLE}\label{sec: olps-solution}

In this section, we introduce our novel approach to adaptive parameter configuration for an online portfolio selection strategy. To address the problem introduced in the previous section, we start by proposing a \emph{parameter configuration map} $f$ that defines a relationship between the compact parameter space $\mathcal{P} \subseteq \mathbb{R}^D$, the space of trading periods $\mathcal{T} \subseteq \mathbb{N}_{+}$ and the space of performance metrics $ \mathcal{M} \subseteq \mathbb{R} $. We define this map as
    \begin{equation}\label{eq: olps-paramter-map}
 				f : \mathcal{P} \times \mathcal{T} \mapsto \mathcal{M},
    \end{equation}
    where parameters $\boldsymbol{ \theta } \in \mathcal{P}$ and performance metrics $m \in  \mathcal{M}$. This idea behind this relationship is predicated on the notion that the sequential non-stationary data induces the temporal evolutions of the best parameters of $g$. This parameter configuration map $f$ captures a temporal-parametric relationship that can be exploited to generate optimal parameters over a horizon of trading periods. Learning the map $f$ would also allow us to gain insights into the behaviour of the underlying trading approach and exploit those insights to adaptively tune $\boldsymbol{\theta}$ for good performance. 

We assume that the parameter configuration map $f$ is of the functional form 
	 \begin{equation}\label{eq: olps-paramter-function}
		f:( \boldsymbol{ \theta }, t ) \mapsto m, 
	 \end{equation}
	where we take the metric $m$ to be the strategy's return at period $t \in \mathcal{T}$ while operating with parameters $\boldsymbol{ \theta }$. We can take $\log f$ be a sample from a zero-mean Gaussian process with covariance function $K$
		\begin{equation}\label{eq: olps-paramter-map-prior}
			\log f \sim \mathcal{GP}(  \boldsymbol{0} , K ).
		\end{equation}  
		We consider $K$ to be a separable spatiotemporal covariance function given by 
			\begin{equation}\label{eq: olps-paramter-map-kernel}
		        K(\{\boldsymbol{ \theta }, t \}, {\{\boldsymbol{ \theta }^{ \prime }, t^ { \prime } \} } ) = 
		        K_{ \text{ P } }( \boldsymbol{ \theta } , \boldsymbol{ \theta }^{ \prime } )
		        \times
		        K_{\text{T}}(t, t^{ \prime } ).
		    \end{equation} 
		    The parameter-space kernel $K_{ \text{P} }$ is taken to be a separable product Rational Quadratic (RQ) kernel  
			     \begin{equation}\label{eq: olps-paramter-map-kernel-space}
				     K_{ \text{P} }( \boldsymbol{\theta}, \boldsymbol{\theta}^{\prime} ) =
				     \sigma_f^{2} \prod_{i=1}^{D} \Bigg( 
				     1 + 
				     \frac{ ( {\theta}_{i} - {\theta}_{i}^{\prime} )^{2} }{  2 \alpha_i l_i^{2} }  
				     \Bigg)^{ - \alpha_i}, 
			    \end{equation}  
			     where the hyper-parameters $ \sigma_f, l_i  , \alpha_i > 0 $. The choice for this kernel was inspired by the fact that we assume the assets that we trade are \emph{liquid} (easily bought or sold), so the effect of any small variations in the proportion of a particular asset on the value of the portfolio will be smooth. Since each asset will have a different characterization of this smoothness, the rational quadratic kernel provides the advantage of being a scale mixture of the squared exponential (SE) kernel with varying length-scales \citep[\S 4.2]{gp1}.
			     	 We take the temporal kernel $K_{ \text{T} }$ to be
				     \begin{equation}\label{eq: olps-paramter-map-kernel-time}
					     K_{ \text{T} }( {t}, {t}^{\prime} ) =
					     \exp{\Bigg(\frac{ ( t - t^{\prime} ) }{ l } \Bigg) } +   
					     \Bigg( 
					     1 + 
					     \frac{ ( {t} - {t}^{\prime} )^{2} }{  2 \alpha l^{2} }  
					     \Bigg)^{ - \alpha},
				     \end{equation}
			     	 where $l_{1}, l_{2}, \alpha > 0 $. This temporal kernel was selected to accommodate both abrupt and smooth temporal changes of varying degrees.

Now that we have a description of the parameter configuration map, the next step is how to use it for adaptively tuning the settings.  To choose the best configuration $ \boldsymbol{ \theta } $ for the online portfolio selection strategy $g$ at every trading step $t$, we are interested in solving  
   \begin{equation}\label{eq: olps-meta}
        \boldsymbol{ \theta }^{ \star }_{t} = \underset{  \boldsymbol{ \theta } \in \mathcal{S}_t  }{ \text{ arg max } } f( \boldsymbol{ \theta }, t ),
    \end{equation}  
    where the parameter configuration map $f( \boldsymbol{ \theta }, t )$ is the meta objective function of the OLPS strategy $g$.

To keep track of the best parameters using the meta objective function, we resort to adaptive Bayesian optimization which will allow us to choose \emph{what} parameters to use in every trading period as the strategy $g$ encounters changing market conditions. We make use of the simplified version of ABO where the times of interest are known. This is because we know the trading frequency in this problem \emph{a priori}. We intelligently evaluate the learnt Gaussian process model using smart search heuristics, which are encoded by the acquisition function $a( \cdot )$, that facilitate the selection of the best parameters to use in the relevant trading periods. The resulting parametric oracle is highlighted in red within in Algorithm \ref{alg: olps_ours}, which shows an updated OLPS procedure.
	\renewcommand{\algorithmicrequire}{\textbf{Input:}}
	\renewcommand{\algorithmicensure}{\textbf{Output:}}
	\begin{algorithm}[htb]
		\caption{OLPS with the Configuration Oracle}
		\label{alg: olps_ours}
		\begin{algorithmic}[1]
			
			\Require $T$: number of trading periods in trading horizon
			\Require $g( \,\cdot \,; \boldsymbol{ \theta } )$: OLPS strategy with parameters $\boldsymbol{ \theta }$
			\Require $m$: performance metric
			\Require GP Prior $\mathcal{GP}(\mu, K_{\text{P}} \odot K_{\text{T}} )$ for $f: (\boldsymbol{ \theta }, t ) \mapsto m$
				
			\Ensure  $\{ \boldsymbol{ \theta }_1, ... , \boldsymbol{ \theta }_t \}$ trace of parameters, where $t \leq T$
			\Ensure  $\{ m_1, ... , m_t \}$ trace of performance metrics, where $t \leq T$
				
			\For {$t = 1,2, ..., T$}
				\State \textcolor{red}{
				Train GP model $\mathcal{GP}(\mu, K)$
				}
				\State \textcolor{red}{ Set bounds of $\mathcal{S}_{t}$ using information from $K_{\mathcal{T}}$}
				\State \textcolor{red}{Select parameters $\boldsymbol{ \theta }^{\star}_{t}$ from
					$$
						\boldsymbol{ \theta }^{\star}_{t} = \underset{ \boldsymbol{ \theta }_{t} \: \in \mathcal{S}_{t} }{ \text{ arg max } }\; a( \boldsymbol{ \theta }, t )
					$$
				}
				\State \textcolor{blue}{
				Strategy learns portfolio $\mathbi{w}_t^{\star}$ by
					$$
						\mathbi{w}_{t }^{  \star  } = \text{ arg max } \; g( \mathbi{w}_{t} ; \mathcal{D},  \boldsymbol{ \theta }^{\star}_{t} )
					$$
					}
				\State Market reveals asset returns
				\State Portfolio $\mathbi{w}_{t }^{  \star  }$ incurs period profits or losses
				\State Strategy updates portfolio selection model
				\State \textcolor{red}{
				Update data for GP model
				} 
			\EndFor
		\end{algorithmic}
	\end{algorithm}
	
\subsection{STRATEGIC INSIGHTS}

The most critical part of the adaptive configuration oracle is the parameter configuration map $f$. The importance of $f$ is more ubiquitous than the configuration task as it generates insights about the OLPS problem in terms of the modelled parameters. Learning $f$ leverages observable information and uncovers intricate patterns for a wide variety of OLPS strategies. What the oracle does in its configuration task is to use these insights to propose the best parameters as time evolving market conditions are revealed to the OLPS strategy.

Using performance metrics as the output space of the configuration map is crucial for uncovering insights from the data.  The performance metrics operate as an optimality criterion that discriminates good parameters from bad while encoding user-specific objectives. Many practitioners have excellent prior knowledge about the settings for their OLPS methods based on their experience of using them. This knowledge will usually include the optimal range of values for good performance and scenarios when particular configurations work excellently. All this information can be incorporated into the oracle via the specification of an appropriate Gaussian process prior and by setting constraints on the non-temporal variables during the acquisition function optimization.

The temporal length-scales provide additional insights on whether an OLPS approach would receive performance gains from the adaptive tuning of its parameters. If the learnt temporal length-scale is large when compared to the horizon of trading periods under consideration, then there is no time variation in the best parameters. On the other hand, if the temporal length-scale is small, then the best parameters evolve with time.

Since BO explores and exploits $f$ to search for the best configurations, sub-optimal performance may be incurred in some exploration steps which can be costly in the OLPS setting. However, there is a remedy for this. The practitioner-based prior art on OLPS configuration becomes useful in introducing constraints for the search that leads to less risky performance.

\section{EXPERIMENTS}\label{sec: olps-experiments}

In this section, we present an extensive set of empirical analyses of our configuration oracle and compare it with the `best' static settings of some popular online portfolio selection methods. For all the experiments, the portfolios are rebalanced on a daily basis. At the end of each trading day, we reconfigure the OLPS methods, and they calculate the target portfolio for the following market opening. The true market data is only revealed to the OLPS algorithms after the relevant portfolios have been constructed, so all the adaptive parameter configurations and portfolio calculations are out-of-sample. The oracle operates by picking daily configurations that aimed at maximizing the daily return of the generated portfolio. We first describe the experimental protocols, the nature of the OLPS algorithms that we configure and then devolve into the results.

\textbf{Comparison Measures:}
In our experiments, we compare various OLPS methods with their best settings as prescribed by their respective authors and variants adaptively tuned by our oracle. We compare the performance of the methods under criteria that fall into the following three classes: (i) absolute returns: how much money the strategy makes over a pre-defined time horizon; (ii) risk: how much financial risk the strategy induces over that time horizon; and (iii) risk-adjusted returns: how profitable the the strategy is when risk is taken into account.

The specific measures under these classes are shown in Table \ref{t: olps-measures}.  We assume there are no transaction costs for all our experiments because we are only interested in gauging the performance improvements resulting from using our configuration oracle.
	\begin{table*}[htp]
		\caption{Comparison measures used in the experiments and their classes.}
		\label{t: olps-measures}
		\begin{center}
			\begin{tabular}{ l l l } 
				{Criteria}  & \multicolumn{2}{c}{Measures} \\
				\midrule
				{Absolute return} & {Cumulative wealth} & {Annualized percentage yield} \\ %
				{Risk}                  & {Annualized standard deviation} & {Maximum drawdown}  \\
				{Risk-adjusted return}     & {Annualized Sharpe ratio} & {Calmer ratio} \\
				\bottomrule
			\end{tabular}	
		\end{center}
	\end{table*}
	
We briefly define the measures shown in Table \ref{t: olps-measures} below:
    \begin{itemize}
        \item Cumulative wealth: This is the total wealth a strategy accumulates assuming an initial investment of \$1. This is the most common measure for comparing different trading strategies and we use it as our main comparator. In general, higher values of cumulative wealth indicate better trading algorithms. Since portfolio returns are multiplicative, the total wealth represents the multiple or percentage increment to the original wealth. This makes cumulative wealth interpretable. 
        \item Annualized percentage yield:  This is the annual rate of return that accounts for compounding interest. As with total wealth, higher values indicate better performance.
        \item Annualized standard deviation: This is the standard deviation of the portfolio returns multiplied by the square root of the number of trading periods in one year, and it is commonly used as a measure of risk. Since it has the same units as the returns, the annualized standard deviation is commonly represented as a percentage. It is also called the \emph{volatility risk}. Unlike cumulative wealth, lower values indicate better risk management.
        \item Maximum drawdown: A drawdown is a difference between the maximum and minimum monetary values of a portfolio during an investment period. The maximum drawdown is the largest difference in portfolio values over a time horizon. It is used to measure the downside risk of a trading strategy.  As with volatility risk, it is usually represented as a percentage, and lower values indicate good performance. 
        \item Annualized Sharpe ratio: This is the average return earned more than the risk-free rate per unit of total risk. This is a popular measure for risk-adjusted performance. In general, as with total wealth, higher values of the Sharpe ratio indicate better trading algorithms.
        \item Calmer ratio: This measures a strategy's performance relative to its risk. It is calculated by dividing the average annual rate of return by the maximum drawdown. It is also called the \emph{drawdown ratio}. As with Sharpe ratio, higher values indicate better risk-adjusted performance.
    \end{itemize}
	
\textbf{Tested Strategies:}
Table \ref{t: olps-strategies} shows the OLPS strategies that were used to test the efficacy of the adaptive configuration oracle. Table \ref{t: olps-strategies} also shows the abbreviations we use for the approaches. We add a suffix `-O' for the version of the methods that are adaptively tuned by our oracle.  Additionally, Table \ref{t: olps-classifications} also shows the classes of trading ideas that the strategies belong to and includes the number of available tunable parameters configured during the experiments. 
	\begin{table*}[t]
		\caption{Trading strategies used for experiments and their representative references.}
		\label{t: olps-strategies}
		\begin{center}
			\begin{tabular}{ l l } 				
				{Algorithm}  & {Reference} \\
				\midrule
				{Market}       & { -- } \\
				{Best-stock (BS)}             & { -- }\\
				{Best Constant Rebalanced Portfolios (BCRP)}  & \citep{fin14}\\
				{Exponential Gradient (EG)}  & \citep{fin22}\\
				{Online Newton Step (ONS)}  & \citep{fin21}\\
				{Passive Aggressive Mean Reversion (PAMR)}  & \citep{LZH+12}\\
				{Confidence-Weighted Mean Reversion (CWMR)}  & \citep{LHZ+13}\\
				{Online Moving Average Reversion (OLMAR)}  & \citep{LH12}\\  
				\bottomrule
			\end{tabular}	
		\end{center}

	\end{table*} 
	
	\begin{table}[t]
		\caption{Classifications and the number tunable of parameters for the trading strategies.}
		\label{t: olps-classifications}
		\begin{center}
			\begin{tabular}{ l l c } 
				
				{Classification} & {Algorithm} & {No. of Params } \\
				\midrule
				{Benchmarks} & {Market}  & { 0 } \\
				{ } & {BS} & { 0 }  \\
				{ } & {BCRP} & { 0 }  \\ 
				\midrule
				{Momentum} & {EG} & { 1 }  \\
				{ } & {ONS} & { 2 }  \\			
				\midrule
				{Mean reversion} & {CWMR}  & { 1 }  \\ 
				{ } & {OLMAR}  & { 2 }    \\ 
				{ } & {PAMR} & { 1 }  \\	
				\bottomrule
			\end{tabular}
		\end{center}
	\end{table}   

\textbf{Datasets:} 
For our empirical analyses, we use publicly available historical daily prices of stock and index data from diverse markets in different time periods. The data used in the experiments is shown in Table \ref{t: olps-datasets}. These datasets were collected by authors of the various OLPS methods, including some of authors of the methods that we test in this paper, and are commonly used for experiments in the OLPS literature. For more details on the datasets, their sources, selection criteria and histories of their use, see \cite{DBOLPS} and \cite{fin15}. 
	\begin{table}[h]
		\caption{Summary of four real market datasets.}
		\label{t: olps-datasets}
		\begin{center}
			\begin{tabular}{ l c c c } 
					{Dataset}  & {\# Assets} & {\# Days} & {Time Frame} \\
					\midrule
					{DJIA}   & {30}  & {507} & {Jan/2001 - Jan/2013} \\ 
					{SP500} & {25} & {1276}  & {Jan/1998 - Jan/2003} \\ 
					{TSE} & {88} & {1258}  & {Jan/1994 - Dec/1998} \\ 
					{MSCI}  & {24}  & {1043}  & {Apr/2006 - Mar/2010} \\ 
					\bottomrule
			\end{tabular}
		\end{center}
	\end{table}
	
\textbf{Implementation Details:}
The OLPS toolbox \citep{li2015olps} was used for implementations of the adaptively configured strategies used for the experiments. Bayesian optimization with the UCB acquisition function was incorporated into the configuration steps within the toolbox. Acquisition function optimization was performed using Particle Swarm Optimization (PSO). The Gaussian process priors used were those specified in Section \ref{sec: olps-solution} and trained using \emph{maximum a posteriori probability} (MAP) estimation with log-normal hyper-priors on all parameter space kernel hyper-parameters, and Gamma hyper-priors on the temporal space kernel hyper-parameters.

The parameters for the first ten trading periods were randomly generated using space-filling latin hypercube sampling with bounds defined by the parameter range constraints. To deal with computational complexity of the GP as the number of data points increases, for all experiments, we used a maximum moving data window of size $300$ of the most recent data points.

All the parameters for the standard versions of the OLPS methods were set according to their original empirical studies whose references are stated in Table \ref{t: olps-strategies}. 
	
\begin{figure}
		\includegraphics[width=\columnwidth]{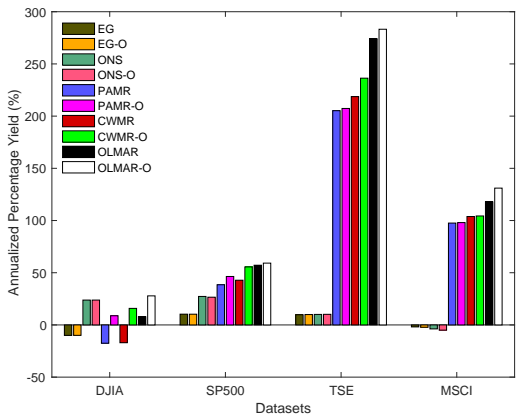}
		\caption{Annualized Percentage Yield.}
		\label{fig: olps-apy}	
	\end{figure}

\textbf{Absolute Returns:}
Table \ref{t: olps-cw} illustrates the main results for cumulative wealth on the four datasets achieved by the standard and adaptively tuned OLPS approaches. The results show that the adaptively-tuned methods achieved better performance than their standard counterparts on most of the tests. Figure \ref{fig: olps-apy} also shows the annualized percentage yields of the approaches and shows the same trend.
 	\begin{table}[h]
		\caption{Cumulative wealth for different OLPS methods.}
		\label{t: olps-cw}
		\begin{center}
			\begin{tabular}{l c c c c} 
				 	
						
				{Methods} &
				
				{DJIA} &
				{SP500} &
				{TSE} &
				{MSCI} \\
				\midrule				
				
				{Market} &
				{0.76} & {1.34} &  {1.61} & {0.91} \\ 
				
				{BS} &  
				{1.19} & {3.78} &  {6.28} & {1.50} \\ 
				
				{BCRP} & 
				{1.24} & {4.04} & {6.78} & {1.51} \\ 
				
				\midrule
								
				{EG}    & 
				{0.81}  & {1.63}  &  {1.59} & {0.93} \\ 
				
				{EG-O}      & 
				{0.81}  & {1.63} & {1.61} & {0.92} \\ 
				
				{ }      & 
				{  } & {  }  & {  } & {  } \\ 
				
				{ONS}   & 
				{1.53} &  {3.34}  & {1.61} & {0.86} \\ 
				
				{ONS-O}      & 
				{1.53}  & {3.25} & {1.62} & {0.99} \\ 
				
				\midrule
				
				{PAMR}        & 
				{0.68}  & {5.09} & {264.86} & {15.23} \\ 
				
				{PAMR-O}        &                                              
				{1.18} & {6.73}  & {274.24} & {15.37} \\ 
				
				{ }      & 
				{  } & {  }  & {  } & {  } \\ 
				
				{CWMR}      & 
				{0.68}  & {5.90} & {332.62} & {17.28} \\ 
				
				{CWMR-O}      &                                                             
				{1.34} & {9.12}  & {357.24} & {17.44} \\ 
				
				{ }      & 
				{  } & {  }  & {  } & {  } \\ 
				
				{OLMAR}      & 
				{1.20}  & {8.63} &  {678.44} & {22.51} \\ 
				
				{OLMAR-O}      &                                 
				{1.63} & {9.59}  & {714.36} & {28.49} \\ 
					
				\bottomrule
						 
			\end{tabular}	
		\end{center}

	\end{table}
	
For the few adaptively-tuned OLPS methods that performed below their standard counterparts, we observed that after training their models, the time dimension lengthscales were longer than their corresponding trading time horizons. This meant that the was no time-varying behaviour learnt by the GP model of the parameter configuration map. Consequently, the exploration steps induced by adaptive Bayesian optimization lead to suboptimal returns in some trading periods which in turn caused the lower overall performance.

Additionally, all the tested methods did not perform well on the DJIA dataset relative to performances on the other datasets. This was because the characteristics inherent to the DJIA dataset not closely match the assumptions made by the tested strategies. The relatively poorer performance on the DJIA dataset is also reflected in the smaller improvements due to our oracle on tests involving the dataset.
	
The results also show that the degree to which adaptive tuning affects the total wealth produced by a strategy on a particular dataset is proportional to how much wealth is generated by the standard version of that strategy relative to the other methods. This behaviour is particularly prevalent when we compare the relative improvements of the momentum-based algorithms (EG and ONS) to those gained by the mean reversion methods (CWMR, PAMR and OLMAR), whose improvements are much greater. 

Table \ref{t: olps-ttests} lists the $t$-test statistics of the better-performing adaptively-tuned momentum-based algorithms.
We use the method for measuring the $t$-statistics of achieving above-market performance as described in \cite{grinold2000active}. This test identifies the \emph{active returns} that are due to the skill of the adaptively-tuned methods and excludes any return that is a function of the market's movement. Therefore, the $t$-statistic provides a test of the statistical significance of the strategy's above-market performance being due to the skill of the algorithm. As rule of thumb, a $t$-statistic of two or more indicates that the performance of the OLPS strategy is due to skill rather than luck. This means that the probability of observing such an active return by luck is only 5\%.
The statistics in Table \ref{t: olps-ttests} validate the improvements via our oracle for those methods and show that they were very less likely due to luck, whose probabilities are at most 8.68\% on the non-comporting DJIA dataset.  Therefore, the improvements due to our oracle occurred with high confidence.

	\begin{table*}
		\caption{Statistical t-test results of the performance by the better-performing methods configured by the oracle.}		
		\label{t: olps-ttests}
		\begin{center}
			\begin{tabular}{ l l c c c c } 
				{Method}     & 
				{Stat. Attr.}  & 
				{DJIA}         &
				{SP500}      &
				{TSE}          &
				{MSCI} \\
				\midrule
				{PAMR-O} & {$t$-statistics} & {1.3628} & {2.3966} & {2.3966} & {6.1694} \\
				{         } & { $p$-value} & {0.0868} & {0.0083} & {0.0083} & {0.0000}\\
				\midrule
				{CWMR-O} & {$t$-statistics} & {1.6768} & {2.7228} & {3.9580} & {6.4074}\\
				{         } & { $p$-value} & {0.0471} & {0.0033} & {0.0000} & {0.0000} \\
				\midrule
				{OLMAR-O} & {$t$-statistics} & {1.8263} & {2.4599} & {3.9305} & {6.3824}  \\
				{          } & {$p$-value}     & {0.0342} & {0.0070} & {0.0000} & {0.0000} \\
				\bottomrule
			\end{tabular}				
		\end{center}
	\end{table*}

\begin{figure}
		\includegraphics[width = .5\textwidth]{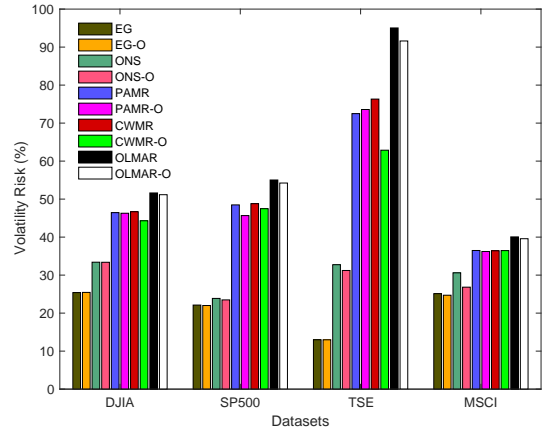}
		\caption{Volatility Risk.}
		\label{fig: olps-vr}
	\end{figure}

\begin{figure}
		\includegraphics[width = .5\textwidth]{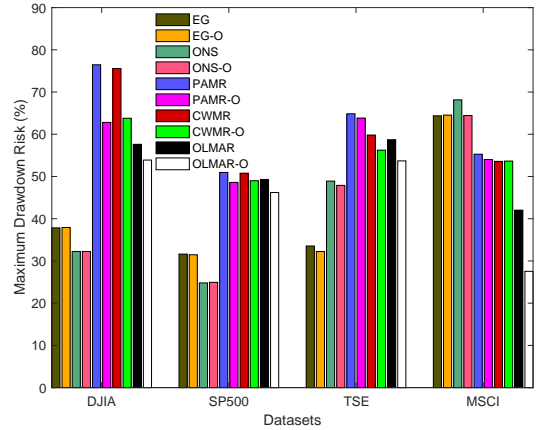}
		\caption{Drawdown Risk.}
		\label{fig: olps-mdd}
	\end{figure}

\textbf{Risk and Risk-Adjusted Returns:}	
An evaluation of the volatility and drawdown risks for the benchmarks and OLPS approaches are shown in Figures \ref{fig: olps-vr} and \ref{fig: olps-mdd}.
	
\begin{figure}
		\includegraphics[width = .5\textwidth]{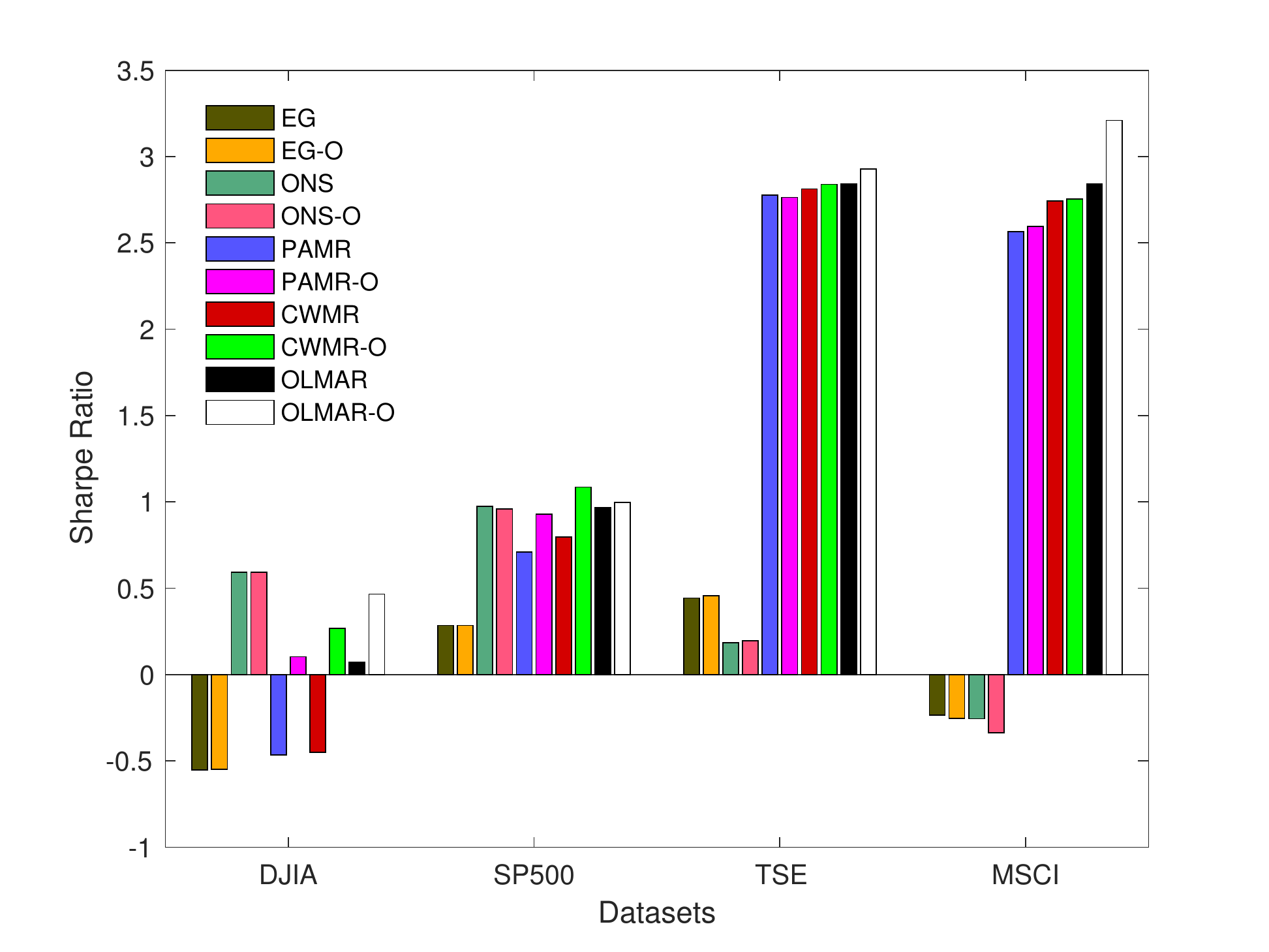} 
		\caption{Annualized Sharpe Ratio.}
		\label{fig: olps-sr}
	\end{figure}

\begin{figure}
		\includegraphics[width = .5\textwidth]{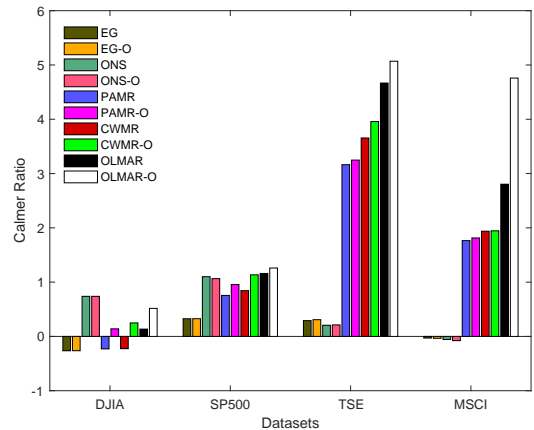}
		\caption{Calmer Ratio.}
		\label{fig: olps-cr}
	\end{figure}

The associated risk-adjusted returns in terms of annualized Sharpe and Calmer ratios are also shown in Figures \ref{fig: olps-sr} and \ref{fig: olps-cr}. 
The results show that the higher the cumulative wealth, the higher the associated risk. This outcome is expected as higher rewards, which our oracle maximized for, are always associated with higher risk. However, other measures that take risk into account can be used for the GP model within the oracle to achieve desired the outcomes within prescribed risk appetites.

\section{DISCUSSION}\label{sec: olps-discussion}

In practice, investment professionals will often have a predefined set of market characteristics that they want to exploit and associated criteria that they want to maximise as they operate within a financial market. This choice of what characteristics an investor will choose to consider will depend on how informative they are about market opportunities. Therefore, parameter configuration maps can be constructed to match these preferences for both the characteristics and performance metrics.

Additionally, one of the benefits of our oracle is that there is a lot of prior art and experience on how existing strategies are configured that can be incorporated into the GP prior of the parameter configuration map. One aspect of the prior art that is commonplace with OLPS strategies is parameter sensitivity analysis which is a time-consuming process that seeks to study how much a strategy's performance changes with different parameters. Our method of adaptive configuration is a natural and fine-tuned extension of the existing scope of parameter sensitivity practices. It provides the benefit of being able to determine if a strategy will benefit from adaptive parameter configuration (or not) as seen in our experiments. Our method is more general and thus subsumes the benefits of parameter sensitivity studies.

\section{CONCLUSION}\label{sec: olps-conclusion}

We described an oracle for adaptively tuning the parameters of OLPS methods. The oracle is supported by a parameter configuration map that is expressed in terms of a spatiotemporal GP prior. The model is intelligently searched using adaptive BO to generate a sequence of parameters for every trading period to maximize the return in that period. We show the efficacy of our oracle by comparing the performance gains of using it on known OLPS approaches operating on a diverse set of financial datasets. Our experimental analyses also showed that the oracle can determine if a method would benefit from adaptive parameter tuning and that it also provides a more general understanding of parameter sensitivity.

\subsubsection*{Acknowledgements}

Favour would like to acknowledge support from the Rhodes Trust and the Oxford-Man Institute of Quantitative Finance. We would also like to thank Yves-Laurent Kom Samo, Marek Musiela, Matthias Qian and Sid Ghoshal for helpful discussions.

\bibliography{bib,olps}
\bibliographystyle{plainnat}

\end{document}